%% file: main.tex
\documentclass{svproc}
\usepackage{url}
\usepackage{type1cm}        
\usepackage{makeidx}         
\usepackage{graphicx}        
\usepackage{multicol}        
\usepackage[bottom]{footmisc}
\usepackage{tikz}
\def\checkmark{\tikz\fill[scale=0.4](0,.35) -- (.25,0) -- (1,.7) -- (.25,.15) -- cycle;} 
\usepackage{enumitem}

\usepackage{subcaption}
\usepackage{newtxtext}       %
\usepackage[varvw]{newtxmath}       
\usepackage{times}

\usepackage[numbers]{natbib}
\usepackage{multicol}
\usepackage[bookmarks=true]{hyperref}
\usepackage{balance} 
\usepackage{amsmath}
\usepackage{multirow}
\usepackage{graphicx}
\usepackage{array}
\usepackage{tabularx}
\usepackage{makecell}
\usepackage{caption}
\usepackage{subcaption}
\usepackage{xcolor}

\newcommand{\ourdataset}{\textbf{CH-MARL}}
\usepackage{tabularx,array,booktabs}

\makeindex             

\begin{document}
\mainmatter              
\title{CH-MARL: A Multimodal Benchmark for Cooperative, Heterogeneous Multi-Agent Reinforcement Learning}
\titlerunning{CH-MARL}  
%
\author{Prasoon Goyal$^{1}$*, Vasu Sharma$^{1}$*, Kaixiang Lin$^{1}$*, Govind Thattai$^{1}$, Qiaozi Gao$^{1}$, Gaurav S. Sukhatme$^{1,2}$}
\institute{$^{1}$Alexa AI, Amazon Inc \hspace{2cm} $^{2}$University of Southern California}

\maketitle              

\begin{abstract}
We propose a multimodal (vision-and-language) benchmark for cooperative and heterogeneous multi-agent learning. We introduce a benchmark multimodal dataset with tasks involving collaboration between multiple simulated heterogeneous robots in a rich multi-room home environment. We provide an integrated learning framework, multimodal implementations of state-of-the-art multi-agent reinforcement learning techniques, and a consistent evaluation protocol. Our experiments investigate the impact of different modalities on multi-agent learning performance. We also introduce a simple message passing method between agents. The results suggest that multimodality introduces unique challenges for cooperative multi-agent learning and there is significant room for advancing multi-agent reinforcement learning methods in such settings.
\end{abstract}
%
\input{./articles/intro}

\input{./articles/related_work}
\input{./articles/benchmark}
\input{./articles/expts}

\input{./articles/conclusion}

%
%
\bibliographystyle{spmpsci}
\bibliography{main}

\end{document}

%% file: articles/intro.tex
\section{Introduction}

We posit that progress in multi-agent learning and its application to multi-robot problems could be sped up with the introduction of standard, sophisticated
environments for training and evaluation. Prior
work on cooperative multi-agent learning has focused on
simplified environments~\cite{lowe2017multi}. 
Visually rich environments that support multi-agent, cooperative tasks
have not been explored until very recently~\cite{wang2021collaborative,jain2020cordial,jain2019two,puig2018virtualhome,tan2020multi}. We propose the first {\em multimodal} benchmark on Cooperative Heterogeneous Multi-Agent Reinforcement Learning (\ourdataset)  wherein two simulated robots must collaboratively find an object and place it at a target location.

\ourdataset\ is built using visually rich scenes from VirtualHome~\cite{puig2018virtualhome}, and includes language. We implement a language generator that procedurally provides feedback to guide embodied agents to achieve tasks. In addition to providing a novel large-scale vision and language dataset for collaborative  task completion in simulated household environments, we conduct  a comprehensive evaluation of several state of the art MARL algorithms under various setting for our collaborative robot benchmark task. 
We investigate and analyze the impact of various aspects of the collaborative MARL algorithms, including heterogeneity and multi-modality. We also propose and implement a message passing interface between agents to enable effective information sharing, especially in decentralized model setups where they would otherwise not have the ability to collaborate with each other.
The results reveal interesting insights: 
\begin{enumerate}
    \item  The multimodal (vision and language) setting presents an extra challenge for existing techniques
    \item Vision and language grounding helps the learning process
    \item Even simple multi-agent communication protocols substantially improve task performance by allowing effective collaboration.
\end{enumerate}
To our knowledge, this is the first dataset to support multiple heterogeneous agents in a virtual environment collaboratively completing a specified task. A comparative study of state of the art embodied AI datasets is in Table \ref{table_datasets}.
We expect this work to contribute towards a standard multi-modal testbed for MARL and foster research in this area. 
\begin{table}[h!]
\centering
\caption{Comparison between state of the art embodied AI simulators and their support for multi agent systems}
\begin{tabular}{c| c | c | c} 
 \hline
 Dataset & Multi agent & Heterogeneous agents  & Language feedback \\ [0.5ex] 
 \hline
 Alfred \cite{alfred} & \checkmark & $\times$ & $\times$ \\ 
 Habitat \cite{habitat}& $\times$ & $\times$ & $\times$ \\
 House3D \cite{house3d} & $\times$ & $\times$ & $\times$ \\
 iGibson \cite{igibson} & \checkmark & $\times$ & $\times$ \\
 Watch and Help \cite{watchandhelp} &  \checkmark & $\times$ & $\times$ \\
 CH-MARL (Ours) &\checkmark & \checkmark & \checkmark\\
\end{tabular}
\label{table_datasets}
\end{table}

%% file: articles/related_work.tex
\section{Related Work}





Collaborative multi-agent RL is well-studied problem. For brevity we only mention the most relevant work here. Kurenkov et al \cite{kurenkov2020semantic} consider the object finding problem, where the target object is described using natural language. However, their focus is to exploit semantic priors about object placements (e.g. cheese is likely to be found in a fridge, which in turn is likely to be in the kitchen). In our setup, objects are not placed according to semantic priors, since our focus is multi-agent collaboration to search for and move objects efficiently. Jain et al (\cite{jain2019two}, \cite{jain2020cordial}) and Nachum et al \cite{nachum2019multi} propose a setting where the agents must perform certain actions synchronously, e.g., for lifting a heavy object. This is different from our setting, which doesn't require synchronous actions; rather, the agents need to communicate with each other to explore the environment efficiently. These prior works assume homogeneous agents, whereas we consider heterogeneous agents. Zhu et al \cite{zhu2021main} study object finding an object in a multi-agent setup, but unlike us, their setting does not involve interactions with the environment, and the agents are homogeneous. Liu et al among others (\cite{liu2020who2com}, \cite{liu2020when2com}, \cite{commreg2018}, \cite{commreg2018_2})) consider the problem of collaborative perception, where there are some degraded agents, and the goal is to learn an efficient communication strategy to improve the observation of the degraded agents. Unlike our setting, their agents are fixed and there is no interaction (navigation or manipulation) with the environment. Baker et al \cite{baker2019emergent} train agents for multi-agent hide-and-seek, where the (homogeneous) agents can navigate and interact in the environment. 

Watch and Help \cite{watchandhelp} by Puig et al. is also a multi agent task completion setup built on top of the Virtual Home \cite{puig2018virtualhome} environment where multiple agents try to complete a human specified task. However, this dataset doesn't have support for heterogeneous agents and doesn't include natural language feedback. IQUAD V1 \cite{iqa} by Gordon et al is built over the AI2-THOR \cite{ai2thor} environment and is a embodied QA dataset with support for multiple agents. The agents are tasked to navigate the environment to answer a question based on it. Again, there is no support for heterogeneous agents or natural language feedback in this dataset. 

%% file: articles/benchmark.tex
\section{Proposed Benchmark}

\subsection{Problem Setting}
\label{sec:problem-setting}

Our setting consists of two simulated robots with different capabilities situated in a home environment. The robots need to cooperate to efficiently complete a task described in natural language. The first robot is a simulated humanoid, while the second is a drone. The humanoid  (H) has an egocentric field of view, and can physically interact with objects in the environment, while the drone  (D) has a top-down view of a part of the environment, but cannot physically interact with objects. Given a task described in natural language, such as ``Put a glass on the desk'', the goal is to complete it in as few steps as possible. In order to complete the task, the robots need to find the objects of interest (i.e. \emph{glass} and \emph{desk}), and the humanoid needs to perform pick-and-place operations to accomplish the desired configuration. Effective inter-robot cooperation is required for efficiency; the drone with its larger field of view can explore the environment more effectively, while only the humanoid can interact with objects. 
Thus, by combining their respective strengths, the two robots can complete the given tasks efficiently.


\noindent \textbf{Environment} We use VirtualHome \cite{puig2018virtualhome}, a Unity-based environment designed for embodied multi-agent collaborative tasks. VirtualHome consists of 7 different scenes; each scene contains multiple rooms. The environment allows initializing objects at different locations in the scene, which can be used to generate various configurations of object placements. We simulate the drone by an overhead camera attached to an invisible agent in VirtualHome. Examples of observations of collaboration between ground agent and drone are shown in Table \ref{tbl:dataset_collaboration}

\noindent \textbf{Tasks} Tasks in the environment involve placing a \emph{graspable} object on a \emph{receptacle} object. There are 45 graspable objects and 16 receptacle objects. For each task, a graspable object, initial receptacle object, and target receptacle object are randomly sampled. The graspable object is initialized at the initial receptacle object, and the goal is to move it to the target receptacle object. The task is described using natural language, such as ``Put the $\langle$graspable object$\rangle$ on the $\langle$receptacle object$\rangle$.'' 

\noindent \textbf{State space} VirtualHome provides both scene graph and visual representations. Hence, our benchmark consists of both --- the scene graph representation that circumvents the object recognition problem (allows focus on multi-agent cooperation), with lower compute requirements, while the visual representation requires object recognition in addition to developing the multi-agent cooperation algorithms, and is closer to the real world.
Each robot receives a local observation at every time step (humanoid: egocentric view of the environment and drone: top-down view of a part of the environment, depending on its current location). In the visual setting, the observations consist of RGB frames for each robot. In the scene graph setting, the observation consists of a graph where the nodes are all the objects present in the visual observation of the robot, and edges describe the relationships between them. For example a coffee mug placed on the dining table will be represented in the scene graph by the nodes "coffee mug" and "table" and the edge between them for the relationship "on".

\noindent \textbf{Action space} To make the setting amenable to reinforcement learning, the action space consists of high-level navigation and manipulation actions, as well as low-level navigation actions. 
\begin{itemize}
    \item High-level navigation actions: These are of the form \texttt{Goto [ROOM]}, where \texttt{ROOM} is one of the rooms in the scene (i.e. kitchen, bedroom, bathroom, livingroom). These actions are available to both the agents.
    \item High-level manipulation actions: Only available to the humanoid robot, and consist of \texttt{Pick} and \texttt{Place} operations. To keep the action space small, we do not require specifying the argument for these actions. Instead, if the robot executes the \texttt{Pick} action when the graspable object of interest is in its view, or the \texttt{Place} action when the receptacle object of interest is in its view (and the robot is holding the graspable object), the actions lead to picking up the target graspable object, and placing the object on the target receptacle object, respectively. Otherwise, the action fails, and results in no change to the environment.
    \item Low-level navigation actions: For the humanoid robot, these actions are \texttt{Move Forward}, \texttt{Turn Left}, and \texttt{Turn Right}, while for the drone, these actions are \texttt{Move Forward}, \texttt{Move Backward}, \texttt{Move Left}, and \texttt{Move Right}.
    \item The Stay action: Both  robots also have a \texttt{Stay} action, which results in no movement of either robot or interaction with the environment.
\end{itemize}
 


\noindent \textbf{Reward} The reward for taking action $a$ at state $s$ is defined in terms of a potential function, $R(s, a, s') = \phi(s') - \phi(s)$, where the potential function $\phi(\cdot)$ is defined as follows:

\[
\phi(s) =
\begin{cases}
 10, & \text{if X placed on Y} \\
 6, & \text{if X grasped, and Y visible to G} \\
 5, & \text{if X grasped, and Y visible to D} \\
 4, & \text{if X grasped, and Y not visible to either G or D} \\
 2, & \text{if X not grasped, and X visible to G} \\
 1, & \text{if X not grasped, and X visible to D} \\
 0, & \text{if X not grasped, and X not visible to either G or D}
\end{cases}
\]
where X is the object of interest, and Y is the target location.


\noindent \textbf{Language Feedback} In addition to the reward, the robots might receive natural language feedback from the environment when they perform a suboptimal action. For instance, if the target object is visible to the humanoid, and it does not pick it up, the feedback may be ``You should have picked up the glass instead of going to the livingroom.''

\subsection{Implementation Details}
\label{sec:dataset}

\noindent \textbf{Trajectory Generation using Planner}
\label{sec:planner}
To create a dataset for offline training, we implement a planner, that given a task, finds a trajectory to complete the task, using privileged information. For instance, if the object of interest is visible to the drone, the humanoid is directed to the location of the object. Using the planner, we generate 6,100 trajectories, which we divide into training, validation, and test splits (\autoref{sec:splits}).

\noindent \textbf{Language Data}
\label{sec:amt}
We generate 100 task descriptions using a single template, and 100 feedback language instructions using 2 templates. We use Amazon Mechanical Turk to obtain 1 paraphrase for each description and feedback item, from which we generate additional templates and extract synonyms for objects. The resulting natural language descriptions and feedback have 183 unique words, and a mean sentence length of 14.04 words. See \autoref{tbl:example-lang} for example task descriptions/feedback.

\begin{table}[h]
\centering
\caption{\textbf{Task descriptions (top) and feedback (bottom).}}
\begin{tabular}{rl} \hline
1.  & Grab the washing scrub and keep it on the kitchencounter                                                      \\
2.  & Place the wine bottle on top of the work table                                                                \\
3.  & Take the lotion and keep it on top of the towel rack                                                          \\
4.  & Pick the coffee pot and put it onto the kitchen counter                                                       \\
5.  & Pick up the mobile phone and place it on top of the sofa                                                      \\ 

\\\hline

1.  & You should have placed the notes on the kitchen table rather than\\ 
& going to the bedroom                         \\
2.  & You didn't have to go to the living room, you should have placed\\ 
& the sports ball on the kitchen counter instead \\
3. & You had to place the board game on the bed instead of moving \\
& forward                                           \\
4. & Rather than turning left, you should have placed the plate on the\\ 
& game box                                    \\
5. & Instead of staying, you should have placed the cooking pot on the\\ 
& kitchen counter                            \\
\hline
\end{tabular}
\label{tbl:example-lang}
\end{table}

\vspace{0.15cm}
\subsection{Evaluation Protocol}

\noindent \textbf{Splits}
\label{sec:splits}
The VirtualHome environment has 7 scenes. The positions of objects can be modified to create different configurations. We create 6000 tasks across scenes 1-5, which are split into 5,500 training, 250 validation-seen and 250 test-seen tasks. 50 tasks are created for scenes 6 and 7 each, which are used as validation-unseen and test-unseen respectively.

\noindent \textbf{Evaluation Metrics}
We compare approaches based on two evaluation metrics -- the success rate of completing tasks, and the episode length for successful completion. These two metrics can be used to compute the path-length-weighted (PLW) score \cite{anderson2018evaluation}  $p_{s} = {s L^*}/{\max(L, L^*)}$ , where $s$ is 1 if the task was successfully completed, and 0 otherwise, $L$ is the number of step taken by the approach to complete the task, and $L^*$ is the optimal number of steps to complete the task, which is estimated as the number of steps in the trajectory generated by the planner (\autoref{sec:planner}). The final score of the algorithm is computed as the average path-length-weighted score across all tasks in the test-unseen split.

\noindent \textbf{Input settings}
We experiment with 2 variants -- scene graph representation and visual representation (\autoref{sec:problem-setting}), and present results for both the settings (\autoref{sec:expts}).


\begin{table}[ht!]
\captionsetup{font=small}
\centering
\caption{Trajectory snippets demonstrating inter-robot collaboration. The task description is "Grab the toy and place it on the bed". In step (a) The ground agent has located the toy while the drone goes and locates the bed. It conveys this information to the drone agent allowing it to quickly navigate to the bed as seen in step (b) and finally in step (c) the ground agent completes the mission by placing the toy on it. }
\begin{tabular}{c | c  c } \hline
Step & Drone view  & ground agent view \\
\hline            
(a) t=0 & \includegraphics[width=0.43\textwidth]{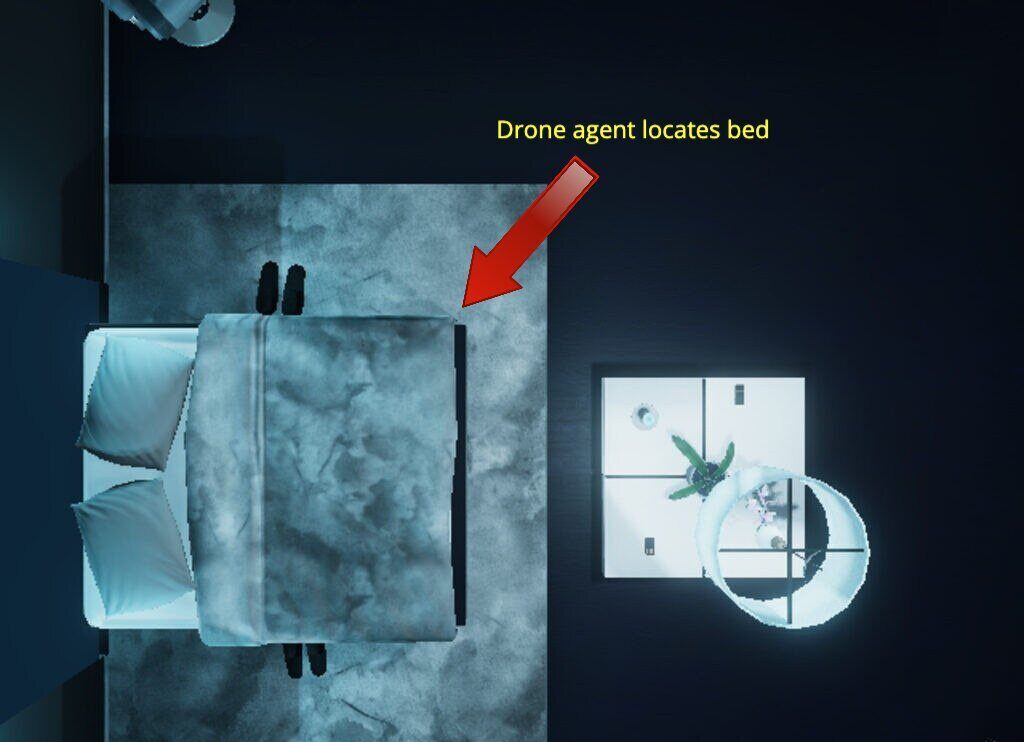} & \includegraphics[width=0.43\textwidth]{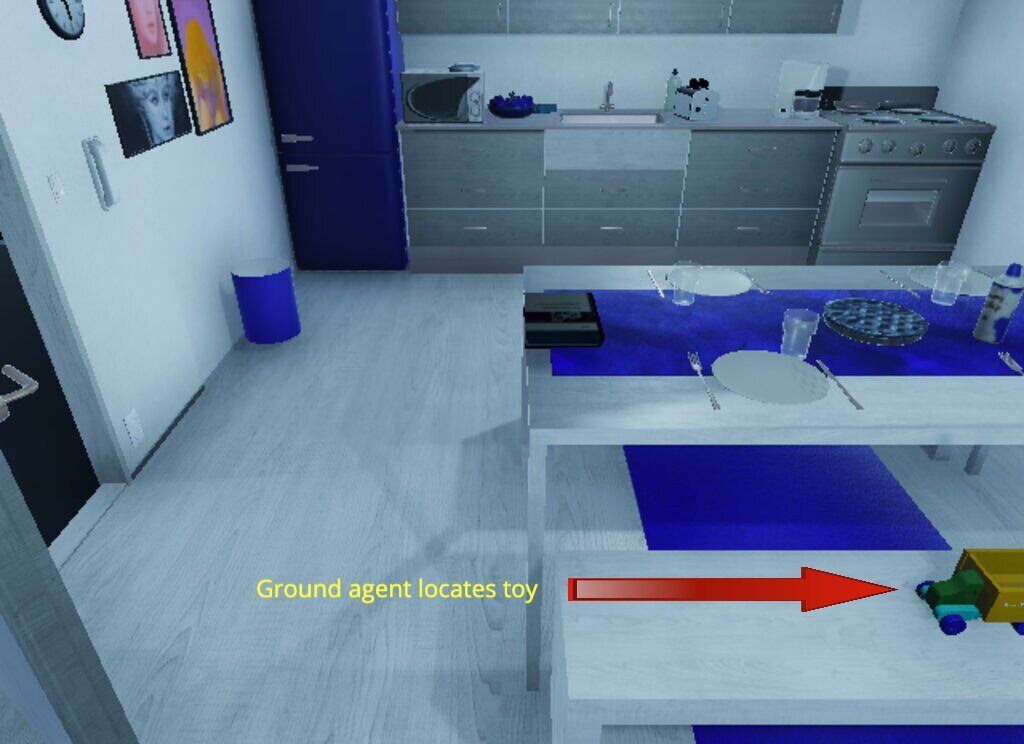} \\
\hline
(b) t=3 & \includegraphics[width=0.43\textwidth]{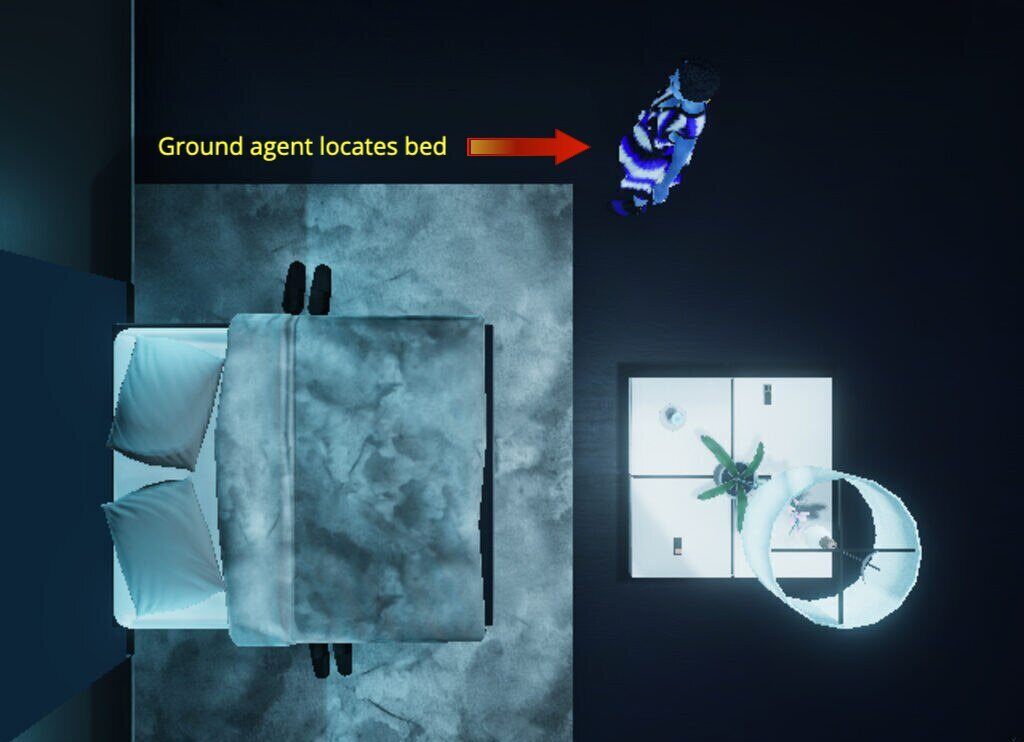} & \includegraphics[width=0.43\textwidth]{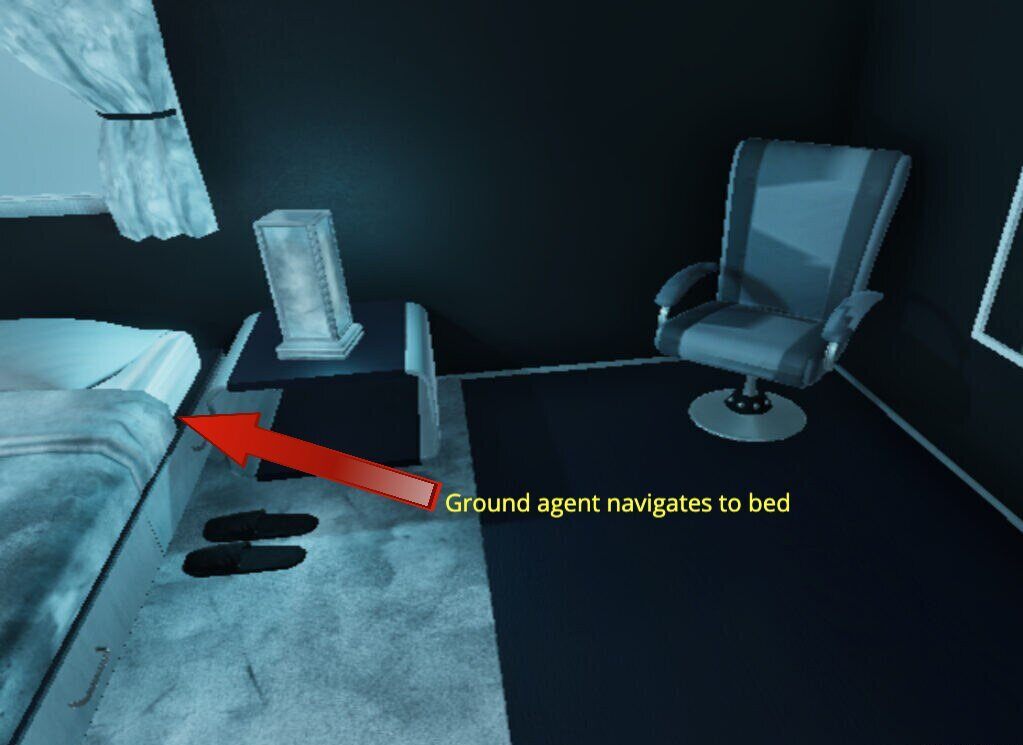} \\
\hline
(c) t=5 & \includegraphics[width=0.43\textwidth]{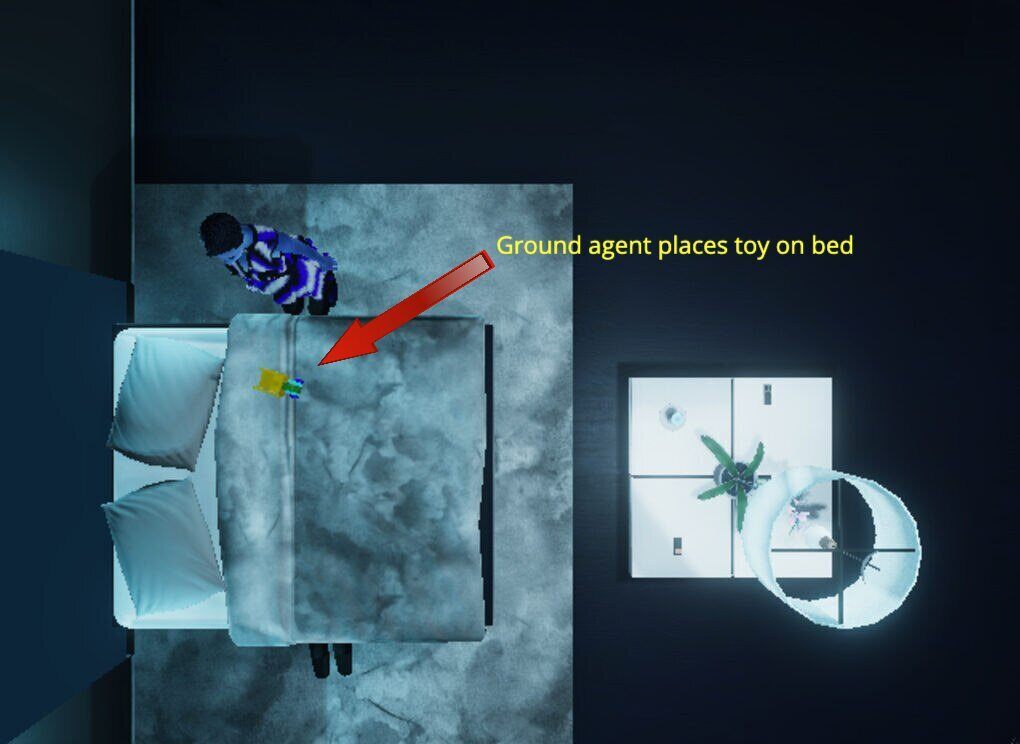} & \includegraphics[width=0.43\textwidth]{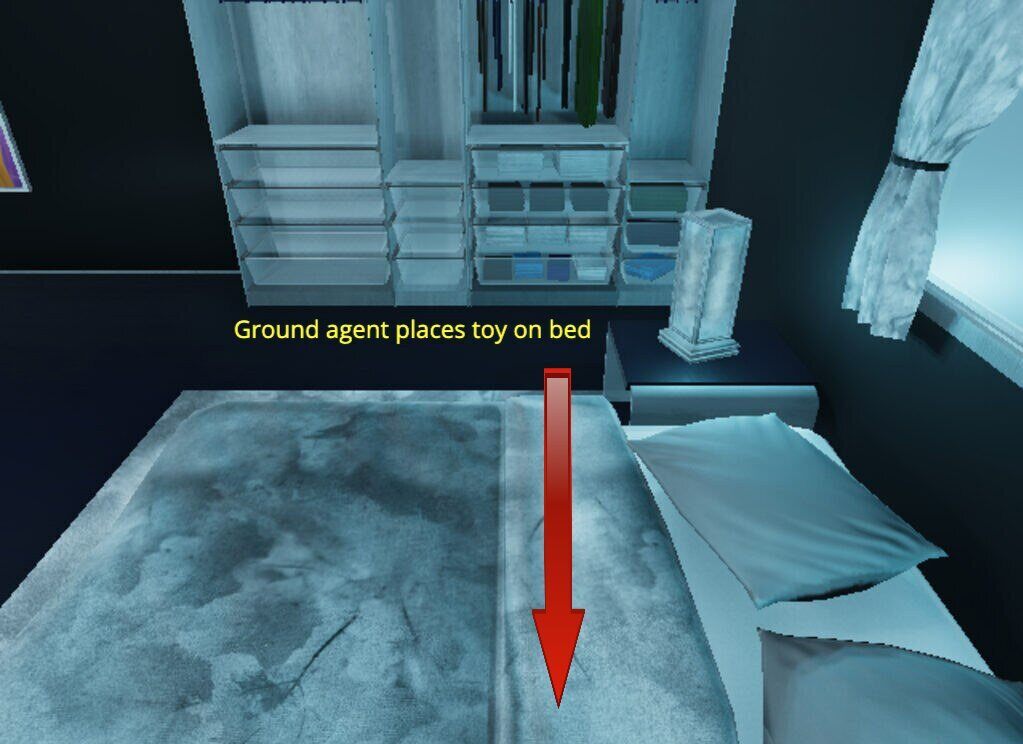} \\
\hline
\end{tabular}
\label{tbl:dataset_collaboration}
\end{table}


\begin{table}[ht!]
\captionsetup{font=small}
\centering
\caption{\textbf{Drone and ground robot view examples from the dataset.} The message vector encodes which room the object of interest or the target receptacle is in, once either robot finds it. The state value encodes progress in the task (higher is better). In the first row, the language instruction is ``Pick up the whipped cream'', the corresponding message is $[0, 1, 0, 0]$, and the state value $\phi(s) = 2$. In the second row the instruction is ``Place it on the bed'', the corresponding message is $[0, 0, 0, 1]$, with state value $\phi(s) = 6$.   }
\begin{tabular}{ c  c } \hline
Drone view  & Ground robot view \\
\hline            
\includegraphics[width=0.45\textwidth]{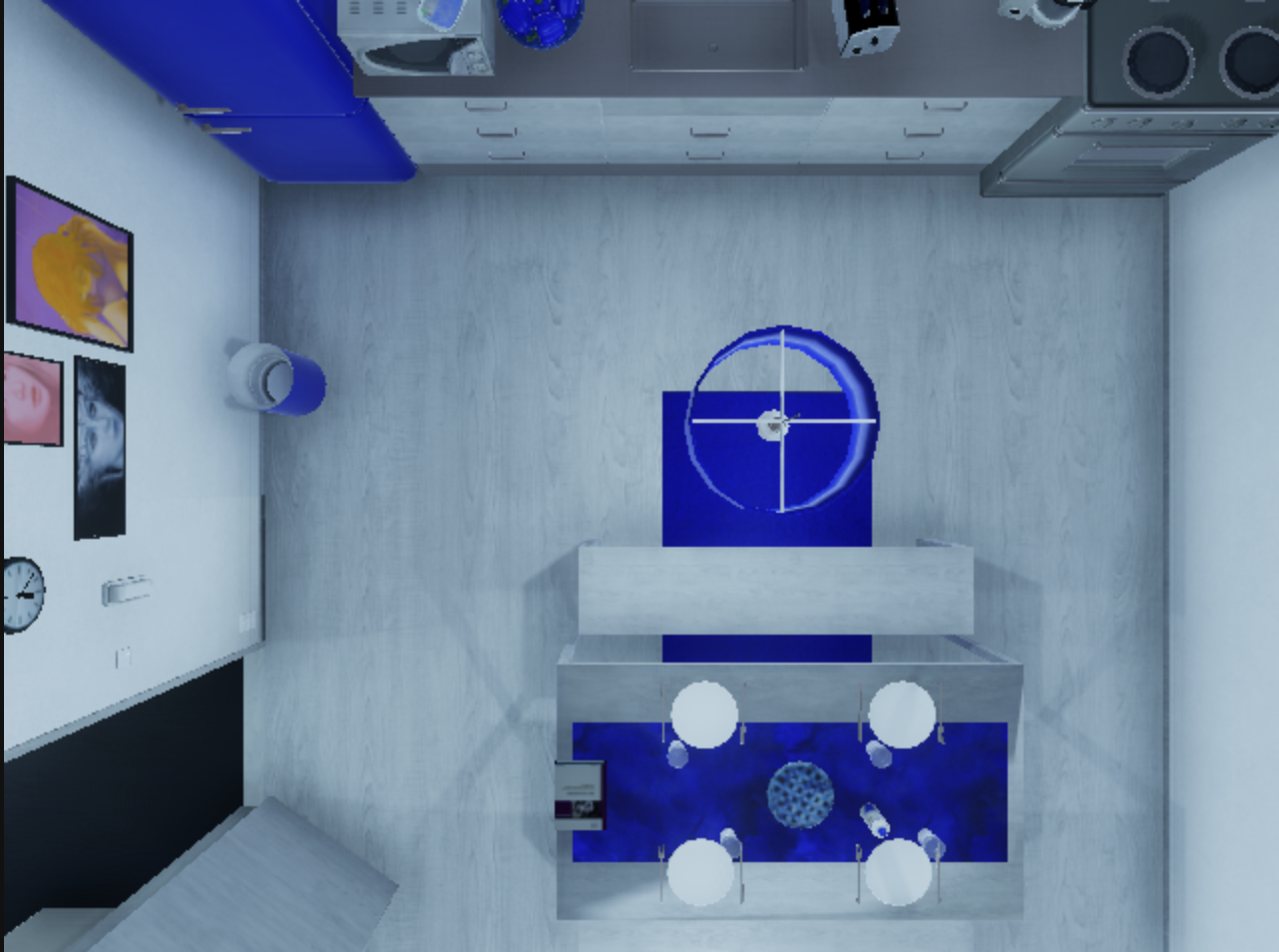} & \includegraphics[width=0.45\textwidth]{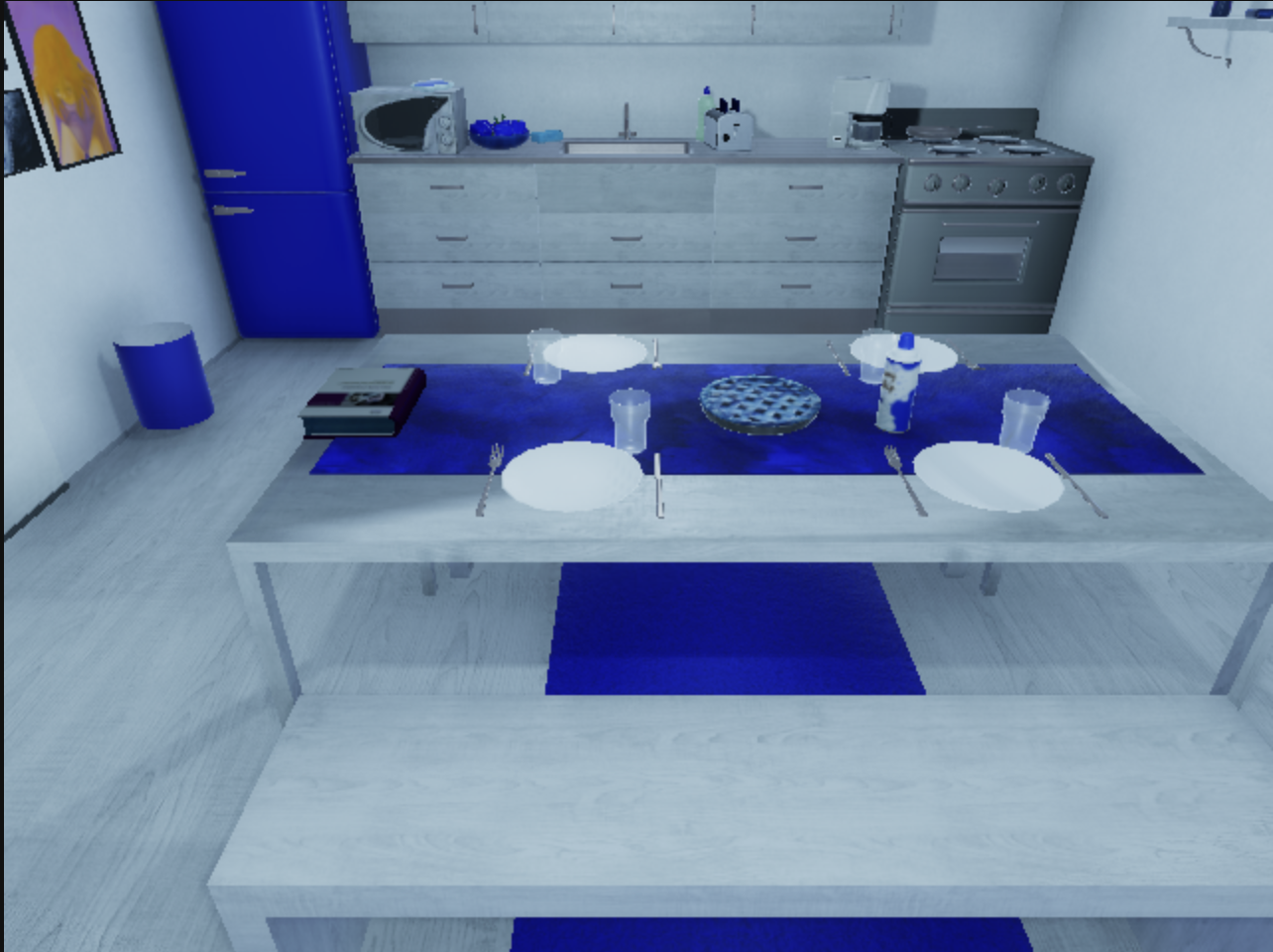} \\
\hline
\includegraphics[width=0.45\textwidth]{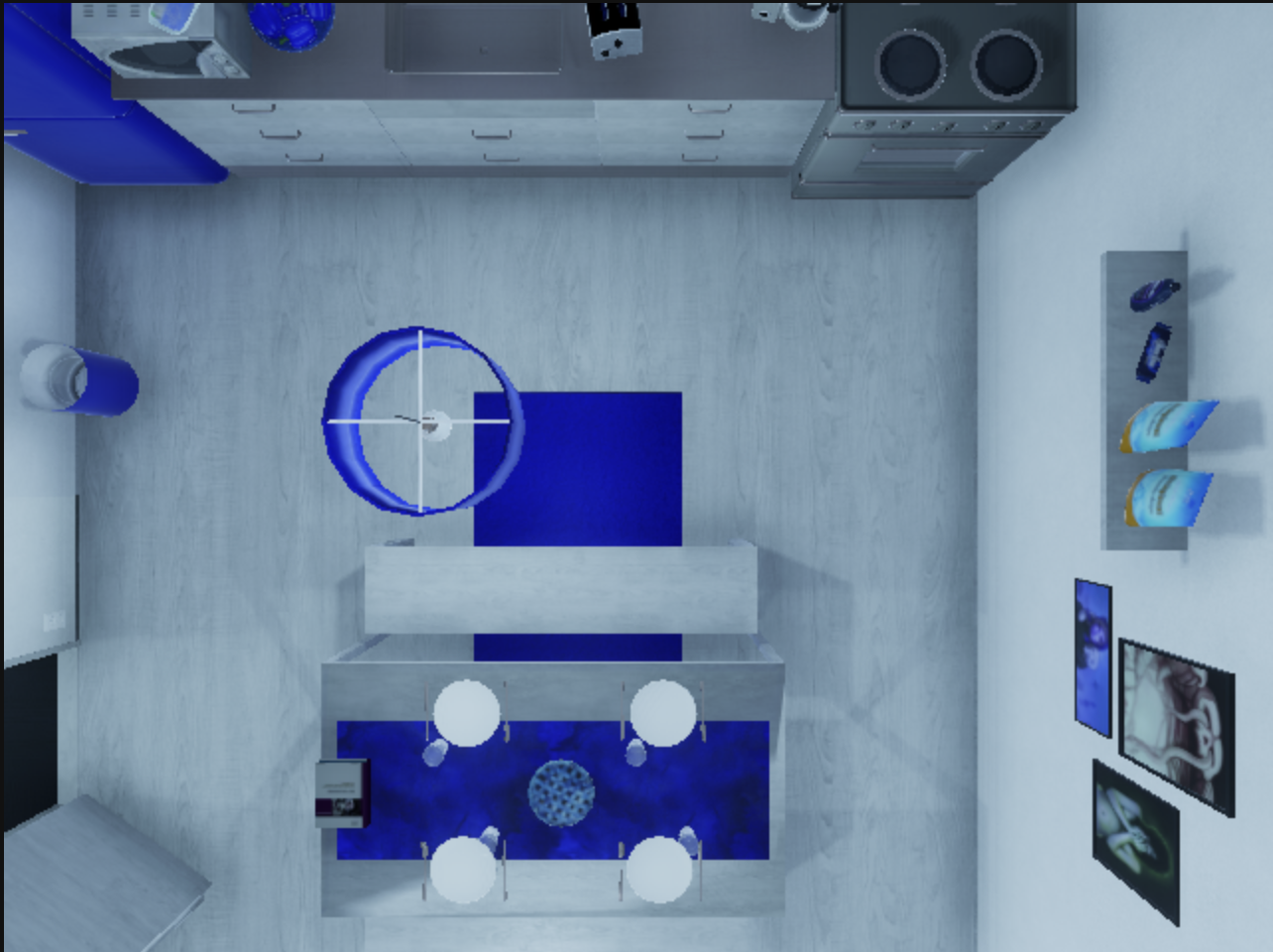} & \includegraphics[width=0.45\textwidth]{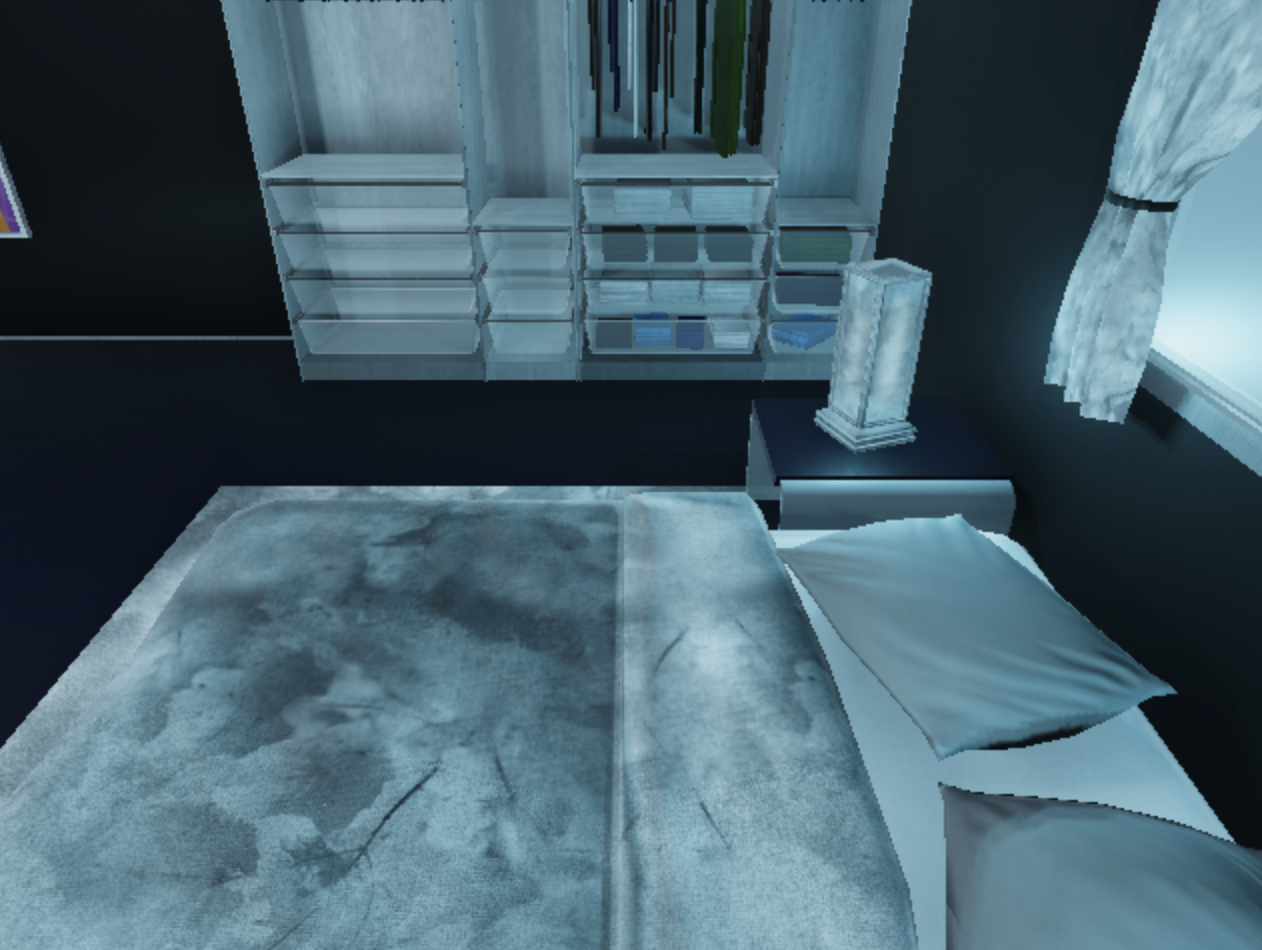} \\
\hline
\end{tabular}
\label{tbl:dataset_examples}
\end{table}

%% file: articles/expts.tex
\section{Experiments}
\label{sec:expts}


\begin{figure}[ht!]
    \centering
    \begin{subfigure}[h]{0.99\textwidth}
        \centering
        \includegraphics[width=\textwidth]{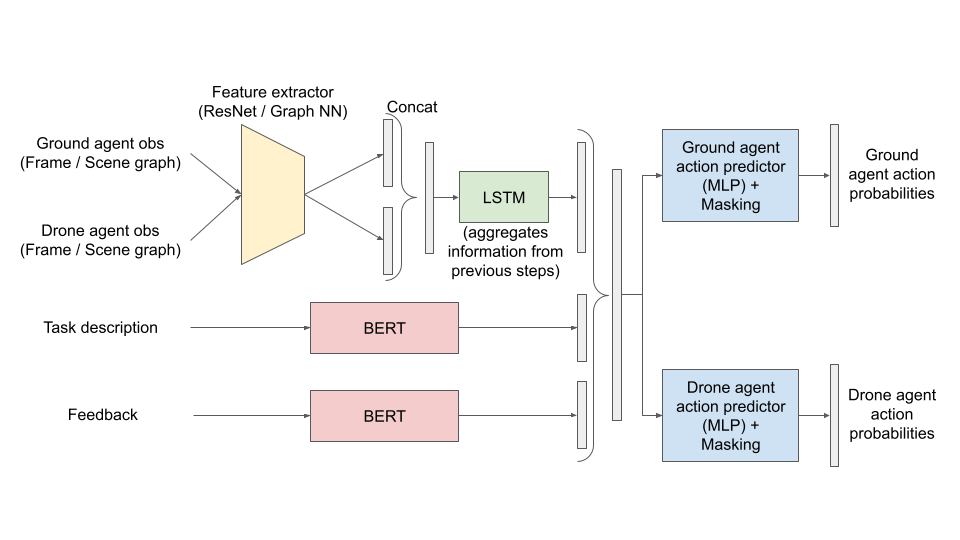}
    \end{subfigure}
    \begin{subfigure}[h]{0.99\textwidth}
        \centering
        \includegraphics[width=\textwidth]{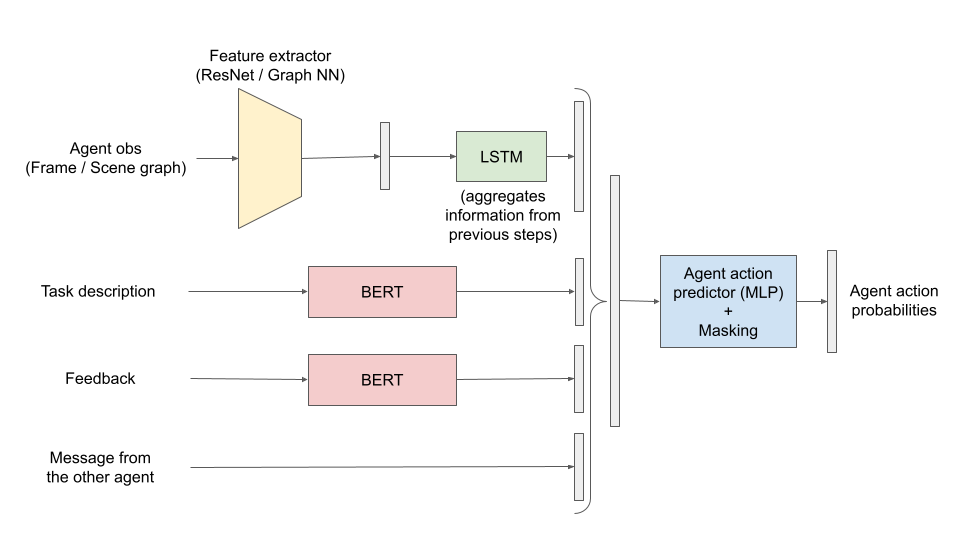}
    \end{subfigure}     
    \caption{
        Model architecture for Centralized (top) and Decentralized (bottom) approaches. As can be seen, the centralized model uses both the ground and drone agent observations as input and generates action predictions for both the agents simultaneously. In the decentralized approach, the model relies only on a single agent's observation while receiving information from the other agent in the form of a message and predicts the action for each agent separately.
    }
    \label{fig:model}
\end{figure}


We benchmark several approaches on our proposed problem setting, including behavior cloning and several state-of-the-art multi-agent RL algorithms. Our model architectures are shown in \autoref{fig:model}. 

\subsection{Feature Extractors}
\label{sec:feature_extractors}

\noindent \textbf{Visual Observation.} The input image is passed through a pretrained ResNet-18 network (\cite{he2016deep}), and the feature vector from the pre-final layer is further projected to a 512-dimensional vector using a linear layer, which is used as the visual representation of the scene.

\noindent \textbf{Scene Graph Observation.} The object class (e.g. bed, table, etc.) and the state (e.g. open, closed, etc.) of each node in the input scene graph is first encoded into vectors using an object class embedding layer and a state embedding layer respectively. These vectors are concatenated to obtain a vector representation for each node. We then apply a Graph Convolution Layer (\cite{kipf2016semi}) to obtain contextualized embeddings for each node, which are aggregated using a mean-pooling operation to obtain the final vector representation of the input scene graph.


\subsection{Message Passing}
\label{sec:message_passing}
A key component of a cooperative multi-agent setup is for agents to communicate effectively. We propose a message passing method which allows  agents to share information with each other. A message is a shared state between the two agents. It is a binary vector of length equal to the number of rooms. Each bit of the vector corresponding to the respective room is set to 1 if either of the agents identify the object of interest to be in that room. We use two such messages, one each for the object of interest and the target receptacle. The messages are used exclusively in the decentralized setups to allow agents to share information. An example simulation of how the agents communicate is shown in Table \ref{tbl:dataset_collaboration}

\begin{table*}[th!]
\centering
\caption{\textbf{Comparison of algorithms with visual observation.} (Ours) here refers to the algorithms with our message passing interface}
\setlength{\tabcolsep}{7pt} 
\begin{tabular}{l|rrrr}
\multicolumn{1}{c}{\multirow{3}{*}{\textbf{Algorithm}}} & \multicolumn{4}{c}{\textbf{Success rate}}                                                                                                                                                                                                                   \\
\multicolumn{1}{c}{}                                    & \multicolumn{2}{c}{\textbf{Validation}}                                 & \multicolumn{2}{c}{\textbf{Test}}                                                                             \\
\multicolumn{1}{c}{}                                    & \multicolumn{1}{c}{\textbf{Seen}} & \multicolumn{1}{c}{\textbf{Unseen}} & \multicolumn{1}{c}{\textbf{Seen}} & \multicolumn{1}{c}{\textbf{Unseen}} \\ \hline
BC; decentralized	&	37.21 $\pm$ 2.40	&	5.36 $\pm$ 1.50	&	42.02 $\pm$ 1.16	&	3.00 $\pm$ 1.00	\\
BC; decentralized(ours)	&	40.89 $\pm$ 2.64	&	6.52 $\pm$ 1.81	&	48.54 $\pm$ 1.66	&	5.02 $\pm$ 0.82	\\
BC; centralized	&	30.80 $\pm$ 2.00	&	4.35 $\pm$ 0.61	&	34.92 $\pm$ 2.96	&	4.00 $\pm$ 0.73\\
\hline
IQL	&	4.05 $\pm$ 0.50	&	2.01 $\pm$ 0.71	&	8.42 $\pm$ 1.12	&	2.33 $\pm$ 0.76		\\
IQL(ours)	&	20.89 $\pm$ 2.01	&	7.56 $\pm$ 0.48	&	22.08 $\pm$ 4.02	&	10.92 $\pm$ 2.11		\\
VDN	&	0.83 $\pm$ 0.11	&	0.67 $\pm$ 0.18	&	2.94 $\pm$ 0.96	&	0.00 $\pm$ 0.00	\\
VDN(ours)	&	14.18 $\pm$ 1.72	&	2.55 $\pm$ 0.40	&	16.56 $\pm$ 2.08	&	5.21 $\pm$ 0.88	\\
QMIX	&	17.73 $\pm$ 2.12	&	3.68 $\pm$ 0.56	&	17.96 $\pm$ 4.58	&	2.33 $\pm$ 0.61	\\
QMIX(ours)	&	22.26 $\pm$ 2.12	&	8.01 $\pm$ 0.56	&	23.85 $\pm$ 4.58	&	12.03 $\pm$ 2.31		\\
QTRAN	&	1.26 $\pm$ 0.23	&	2.01 $\pm$ 0.29	&	2.52 $\pm$ 0.72	&	0.33 $\pm$ 0.08		\\
QTRAN(ours) &	13.69 $\pm$ 1.22	&	7.07 $\pm$ 0.60	&	19.58 $\pm$ 1.66	&	8.00 $\pm$ 0.72	\\

\end{tabular}
\subcaption*{(a)}
\begin{tabular}{l|rrrr}
\multicolumn{1}{c}{\multirow{3}{*}{\textbf{Algorithm}}} & \multicolumn{4}{c}{\textbf{PLW score}}                                                                                                                                                                                                                   \\
\multicolumn{1}{c}{}                                    & \multicolumn{2}{c}{\textbf{Validation}}                                 & \multicolumn{2}{c}{\textbf{Test}}                                                                             \\
\multicolumn{1}{c}{}                                    & \multicolumn{1}{c}{\textbf{Seen}} & \multicolumn{1}{c}{\textbf{Unseen}} & \multicolumn{1}{c}{\textbf{Seen}} & \multicolumn{1}{c}{\textbf{Unseen}} \\ \hline
BC; decentralized	&	30.34 $\pm$ 2.02	&	4.84 $\pm$ 0.78	&	35.43 $\pm$ 1.99	&	2.09 $\pm$ 0.81	\\
BC; decentralized(ours)	&	37.42 $\pm$ 2.51	&	5.51 $\pm$ 0.82	&	40.11 $\pm$ 2.14	&	3.50 $\pm$ 1.31	\\
BC; centralized	&	23.07 $\pm$ 2.55	&	4.19 $\pm$ 0.73	&	27.37 $\pm$ 3.06	&	2.29 $\pm$ 0.74	\\
\hline
IQL	&	2.68 $\pm$ 0.97	&	1.79 $\pm$ 0.51	&	6.31 $\pm$ 1.32	&	1.73 $\pm$ 0.21	\\
IQL(ours)	&	13.80 $\pm$ 2.55	&	5.54 $\pm$ 1.12	&	19.05 $\pm$ 2.55	&	6.10 $\pm$ 1.81	\\
VDN	&		0.60 $\pm$ 0.09	&	0.51 $\pm$ 0.14	&	2.37 $\pm$ 0.81	&	0.00 $\pm$ 0.00	\\
VDN(ours)	& 10.32 $\pm$ 1.14	&	3.76 $\pm$ 0.65	&	12.22 $\pm$ 1.02	&	3.00 $\pm$ 0.28	\\
QMIX	&	14.09 $\pm$ 0.71	&	3.39 $\pm$ 1.06	&	15.21 $\pm$ 2.91	&	1.88 $\pm$ 0.79	\\
QMIX(ours)	& 15.55 $\pm$ 2.71	&	6.34 $\pm$ 1.06	&	19.36 $\pm$ 2.91	&	6.45 $\pm$ 2.29	\\
QTRAN	&	1.09 $\pm$ 0.12	&	1.45 $\pm$ 0.72	&	2.45 $\pm$ 0.31	&	0.33 $\pm$ 0.08	\\
QTRAN(ours) & 9.73 $\pm$ 0.89	&	5.94 $\pm$ 0.61		&	15.30 $\pm$ 1.25	&	4.82 $\pm$ 0.55	\\

\end{tabular}
\subcaption*{(b)}
\label{tbl:results-visual}
\end{table*}

\begin{table*}[th!]
\centering
\caption{\textbf{Comparison of algorithms with scene graph observation.} (Ours) here refers to the algorithms with our message passing interface}
\setlength{\tabcolsep}{7pt} 
\begin{tabular}{l|rrrr}
\multicolumn{1}{c}{\multirow{3}{*}{\textbf{Algorithm}}} & \multicolumn{4}{c}{\textbf{Success rate}}                                                                                                                                                                                                                   \\
\multicolumn{1}{c}{}                                    & \multicolumn{2}{c}{\textbf{Validation}}                                 & \multicolumn{2}{c}{\textbf{Test}}                                                                             \\
\multicolumn{1}{c}{}                                    & \multicolumn{1}{c}{\textbf{Seen}} & \multicolumn{1}{c}{\textbf{Unseen}} & \multicolumn{1}{c}{\textbf{Seen}} & \multicolumn{1}{c}{\textbf{Unseen}} \\ \hline
BC; decentralized	&	41.87 $\pm$ 3.94	&	11.47 $\pm$ 2.66	&	47.19 $\pm$ 0.55	&	13.67 $\pm$ 3.21	\\
BC; decentralized(ours)	&	46.25 $\pm$ 3.94	&	21.08 $\pm$ 2.66	&	53.55 $\pm$ 0.55	&	18.04 $\pm$ 3.21	\\
BC; centralized	&	40.08 $\pm$ 3.81	&	15.15 $\pm$ 1.13	&	33.61 $\pm$ 2.96	&	10.01 $\pm$ 1.93	\\
\hline
IQL	&	14.18 $\pm$ 5.18	&	14.49 $\pm$ 2.59	&	14.53 $\pm$ 5.33	&	13.67 $\pm$ 5.13		\\
IQL(ours)	&	24.16 $\pm$ 3.18	&	23.46 $\pm$ 2.59	&	24.89 $\pm$ 4.33	&	21.26 $\pm$ 4.13	\\
VDN	&	16.85 $\pm$ 1.89	&	17.74 $\pm$ 3.61	&	19.03 $\pm$ 2.08	&	14.67 $\pm$ 4.73		\\
VDN(ours)	&	34.87 $\pm$ 1.89	&	26.04 $\pm$ 3.61	&	36.70 $\pm$ 2.08	&	27.51 $\pm$ 3.73	\\
QMIX	&	20.37 $\pm$ 2.90	&	14.78 $\pm$ 1.67	&	23.48 $\pm$ 2.23	&	18.67 $\pm$ 4.51		\\
QMIX(ours)	&	39.25 $\pm$ 2.90	&	27.85 $\pm$ 1.67	&	38.13 $\pm$ 2.23	&	28.95 $\pm$ 3.51	\\
QTRAN	&	3.04 $\pm$ 1.14	&	2.69 $\pm$ 0.91	&	2.94 $\pm$ 1.75	&	3.00 $\pm$ 0.61	\\
QTRAN(ours)	&	12.42 $\pm$ 2.27	&	10.07 $\pm$ 3.83	&	11.45 $\pm$ 3.03	&	10.05 $\pm$ 2.20\\
\end{tabular}
\subcaption*{(a)}
\begin{tabular}{l|rrrr}
\multicolumn{1}{c}{\multirow{3}{*}{\textbf{Algorithm}}} & \multicolumn{4}{c}{\textbf{PLW Score}}                                                                                                                                                                                                                   \\
\multicolumn{1}{c}{}                                    & \multicolumn{2}{c}{\textbf{Validation}}                                 & \multicolumn{2}{c}{\textbf{Test}}                                                                             \\
\multicolumn{1}{c}{}                                    & \multicolumn{1}{c}{\textbf{Seen}} & \multicolumn{1}{c}{\textbf{Unseen}} & \multicolumn{1}{c}{\textbf{Seen}} & \multicolumn{1}{c}{\textbf{Unseen}} \\ \hline
BC; decentralized		&	35.93 $\pm$ 2.95	&	9.51 $\pm$ 1.80	&	41.78 $\pm$ 1.33	&	11.36 $\pm$ 3.08	\\
BC; decentralized(ours)	&	38.17 $\pm$ 2.95	&	15.06 $\pm$ 1.80	&	45.35 $\pm$ 1.33	&	13.72 $\pm$ 3.08	\\
BC; centralized	&	32.73 $\pm$ 3.60	&	11.89 $\pm$ 0.39	&	26.77 $\pm$ 16.95	&	7.18 $\pm$ 3.11	\\
\hline
IQL	&	9.91 $\pm$ 4.45	&	10.01 $\pm$ 1.91	&10.80 $\pm$ 4.22	&	10.78 $\pm$ 3.12	\\
IQL(ours)		&	19.94 $\pm$ 2.45	&	19.21 $\pm$ 1.91	&19.73 $\pm$ 3.22	&	17.05 $\pm$ 3.12	\\
VDN	&	11.36 $\pm$ 2.06	&	13.76 $\pm$ 2.37	&	13.68 $\pm$ 1.54	&	11.52 $\pm$ 2.87	\\
VDN(ours)	&	23.37 $\pm$ 2.06	&	18.24 $\pm$ 2.37	&	27.86 $\pm$ 1.54	&	17.16 $\pm$ 2.87\\
QMIX	&	13.73 $\pm$ 3.56	&	12.84 $\pm$ 0.58	&	16.71 $\pm$ 3.50	&	14.67 $\pm$ 5.98\\
QMIX(ours) &	25.36 $\pm$ 3.56	&	21.44 $\pm$ 0.58	&	28.68 $\pm$ 3.50	&	20.05 $\pm$ 5.98\\
QTRAN &	2.57 $\pm$ 0.82	&	2.64 $\pm$ 0.55	&	2.57 $\pm$ 1.01	&	2.67 $\pm$ 0.79		\\
QTRAN(ours)	&	10.57 $\pm$ 3.46	&	6.34 $\pm$ 3.74	&	9.12 $\pm$ 4.13	&	7.42 $\pm$ 1.63	\\
\end{tabular}
\subcaption*{(b)}
\label{tbl:results-symbolic}
\end{table*}

\subsection{Algorithms}
\label{sec:algorithms}
\noindent \textbf{Behavior Cloning}
\label{sec:behavior-cloning}
We use the (state, action) pairs in the dataset (\autoref{sec:dataset}) to train policy networks using supervised learning, for both centralized and decentralized scenarios. The decentralized scenario uses message passing between agents. For each scenario, we experiment with both visual and scene graph representations, where the states are encoded using the feature extractor architectures (\autoref{sec:feature_extractors}), and the networks are trained end-to-end using an Adam optimizer.

\noindent \textbf{Decentralized RL Algorithms}
Next, we benchmark several state-of-the-art decentralized RL algorithms, described below.\\
    \noindent \textbf{IQL:} Independent Q-learning trains the Q-function of each agent on its history of local observations.\\
    \noindent \textbf{VDN:} Value Decomposition Network decomposes the joint Q-function into a sum of  Q-functions of the individual agents, and trains each agent on its own Q-function using DQN loss.\\
    \noindent \textbf{QMIX:} extends VDN by relaxing the decomposition of the joint Q-function to be any monotonic function of the individual Q-functions, and trains as in VDN.\\
    \noindent \textbf{QTRAN:} extends both VDN and QMIX by transforming the joint Q-function into an alternate that is expected to be easier to factorize. We base our implementation on \cite{samvelyan19smac}, \cite{foerster2016learning} and extend it for explicit message-passing  (\autoref{sec:feature_extractors}).



\subsection{Results}
\label{sec:results}

Our results (\autoref{tbl:results-visual} and \autoref{tbl:results-symbolic}) show that: 
\begin{enumerate}[label=(\alph*)]
    \item The message passing-based decentralized models (labelled (ours)) are significantly better than their non message-passing based versions. Decentralized behavior-cloning performs best in seen environments in PLW score, both in the visual and scene-graph representations. Both highlight the importance of effective communication 
    \item Each of the RL models perform better on unseen test and validation environments (best PLW $20.05$) compared to behavior cloning methods (best PLW $13.72$), demonstrating their ability to generalize to newer and unseen environments
    \item The visual observation (best test-unseen PLW score: $6.45$) is significantly harder than scene graph observation (best test-unseen PLW score: $20.05$). A potential reason for this could be the need for better representing visual features using a feature extractor trained on in-domain data.
    \item All the existing algorithms achieve a relatively low PLW score on the unseen splits, suggesting  \ourdataset ~could spur creation of new algorithms/models.
\end{enumerate}



\subsection{Analysis of Natural Language Feedback}

To understand whether the language feedback is helpful, we run an ablation experiment where the language feedback is turned off. We choose QMIX as the algorithm and study four different input settings including visual input, visual input with feedback, scene graph input, scene graph input with feedback. The results of this experiment and presented in Table~\ref{tbl:results-ablation}. We can clearly observe that language feedback provides ~10\% relative improvement over the baselines without language feedback for both visual and scene graph inputs demonstrating the necessity of the natural language feedback.

\begin{table*}[]
\centering
\caption{Ablation experiments on QMIX}
\setlength{\tabcolsep}{6pt} 
\begin{tabular}{l|rrrr}
\multicolumn{1}{c}{\multirow{3}{*}{\textbf{Algorithm}}} & \multicolumn{4}{c}{\textbf{Success rate}}                                                                                                                                                                                                                   \\
\multicolumn{1}{c}{}                                    & \multicolumn{2}{c}{\textbf{Validation}}                                 & \multicolumn{2}{c}{\textbf{Test}}                                                                             \\
\multicolumn{1}{c}{}                                    & \multicolumn{1}{c}{\textbf{Seen}} & \multicolumn{1}{c}{\textbf{Unseen}} & \multicolumn{1}{c}{\textbf{Seen}} & \multicolumn{1}{c}{\textbf{Unseen}} \\ \hline
Visual, with feedback	&	22.26 $\pm$ 2.12	&	8.01 $\pm$ 0.56	&	23.85 $\pm$ 4.58	&	12.03 $\pm$ 2.31	\\
Visual; w/o feedback	&	20.47 $\pm$ 2.86	&	7.55 $\pm$ 1.15	&	21.45 $\pm$ 2.07	&	10.33 $\pm$ 0.58	\\ \hline
Scene graph, with feedback	&	39.25 $\pm$ 2.90	&	27.85 $\pm$ 1.67	&	38.13 $\pm$ 2.23	&	28.95 $\pm$ 3.51	\\
Scene graph; w/o feedback	&	36.54 $\pm$ 2.76	&	25.73 $\pm$ 1.43	&	36.23 $\pm$ 3.70	&	15.33 $\pm$ 4.62	\\
\end{tabular}
\subcaption*{(a)}

\begin{tabular}{l|rrrr}
\multicolumn{1}{c}{\multirow{3}{*}{\textbf{Algorithm}}} & \multicolumn{4}{c}{\textbf{PLW score}}                                                                                                                                                                                                                   \\
\multicolumn{1}{c}{}                                    & \multicolumn{2}{c}{\textbf{Validation}}                                 & \multicolumn{2}{c}{\textbf{Test}}                                                                             \\
\multicolumn{1}{c}{}                                    & \multicolumn{1}{c}{\textbf{Seen}} & \multicolumn{1}{c}{\textbf{Unseen}} & \multicolumn{1}{c}{\textbf{Seen}} & \multicolumn{1}{c}{\textbf{Unseen}} \\ \hline
Visual, with feedback		&	15.55 $\pm$ 2.71	&	6.34 $\pm$ 1.06	&	19.36 $\pm$ 2.91	&	6.45 $\pm$ 2.29	\\
Visual; w/o feedback &	14.90 $\pm$ 2.00	&	6.13 $\pm$ 1.02	&	17.31 $\pm$ 0.98	&	6.18 $\pm$ 0.75	\\ \hline
Scene graph, with feedback		&	25.36 $\pm$ 3.56	&	21.44 $\pm$ 0.58	&	28.68 $\pm$ 3.50	&	20.05 $\pm$ 5.98	\\
Scene graph; w/o feedback		&	11.93 $\pm$ 4.04	&	16.80 $\pm$ 1.20	&	12.41 $\pm$ 6.10	&	11.95 $\pm$ 3.36	\\
\end{tabular}
\subcaption*{(b)}
\label{tbl:results-ablation}
\end{table*}

%% file: articles/conclusion.tex
\section{Conclusions}
\label{sec:conc}

We proposed \ourdataset, a new multimodal, multiagent, cooperative learning benchmark, with heterogeneous agents -- a simulated humanoid with an egocentric field of view that can interact with objects in the environment, and a simulated drone with a larger field of view that cannot physically interact with the environment. We create tasks that require effective collaboration between agents. We introduce a simple message passing-based communication interface to allow efficient collaboration between agents, leading to significant performance gains over the non communicative baselines, highlighting the need for better communication between the agents to improve task success. We benchmark existing algorithms on the proposed problem; there is significant room for improvement in a multimodal setup to solve tasks effectively by developing new algorithms that leverage the strengths of each agent, and learn an efficient cooperative policy.

\section{Future Work}
\label{sec:future}
For future work, we plan to extensively study better communication methods between the agents to allow for more effective collaboration between them. We plan to explore both dialog and symbolic communication techniques 
Our setup allows for addition of more robots with different capabilities allowing us to explore how the agent dynamics and communication patterns change as more agents with varying abilities are introduced to the environment to accomplish a task. Our testbed  also enables the exploration of advanced multimodal models which could better leverage the rich multimodal information the environment provides.

%% file: main.bbl
\begin{thebibliography}{10}
\providecommand{\url}[1]{{#1}}
\providecommand{\urlprefix}{URL }
\expandafter\ifx\csname urlstyle\endcsname\relax
  \providecommand{\doi}[1]{DOI~\discretionary{}{}{}#1}\else
  \providecommand{\doi}{DOI~\discretionary{}{}{}\begingroup
  \urlstyle{rm}\Url}\fi

\bibitem{anderson2018evaluation}
Anderson, P., Chang, A., Chaplot, D.S., Dosovitskiy, A., Gupta, S., Koltun, V.,
  Kosecka, J., Malik, J., Mottaghi, R., Savva, M., et~al.: On evaluation of
  embodied navigation agents.
\newblock arXiv preprint arXiv:1807.06757  (2018)

\bibitem{baker2019emergent}
Baker, B., Kanitscheider, I., Markov, T., Wu, Y., Powell, G., McGrew, B.,
  Mordatch, I.: Emergent tool use from multi-agent autocurricula.
\newblock arXiv preprint arXiv:1909.07528  (2019)

\bibitem{foerster2016learning}
Foerster, J.N., Assael, Y.M., De~Freitas, N., Whiteson, S.: Learning to
  communicate with deep multi-agent reinforcement learning.
\newblock arXiv preprint arXiv:1605.06676  (2016)

\bibitem{iqa}
Gordon, D., Kembhavi, A., Rastegari, M., Redmon, J., Fox, D., Farhadi, A.: Iqa:
  Visual question answering in interactive environments.
\newblock In: Computer Vision and Pattern Recognition (CVPR), 2018 IEEE
  Conference on. IEEE (2018)

\bibitem{commreg2018_2}
Gurumurthy, S., Agarwal, A., Sharma, V., Lewis, M., Sycara, K.: Community
  regularization of visually-grounded dialog.
\newblock International Conference on Autonomous Agents and Multiagent
  Systems(AAMAS 2019)  (2019)

\bibitem{commreg2018}
Gurumurthy, S., Agarwal, A., Sharma, V., Sycara, K.P.: Mind your language:
  Learning visually grounded dialog in a multi-agent setting.
\newblock Adaptive Learning Agents (ALA) 2018  (2018)

\bibitem{he2016deep}
He, K., Zhang, X., Ren, S., Sun, J.: Deep residual learning for image
  recognition.
\newblock In: Proceedings of the IEEE conference on computer vision and pattern
  recognition, pp. 770--778 (2016)

\bibitem{jain2020cordial}
Jain, U., Weihs, L., Kolve, E., Farhadi, A., Lazebnik, S., Kembhavi, A.,
  Schwing, A.: A cordial sync: Going beyond marginal policies for multi-agent
  embodied tasks.
\newblock In: European Conference on Computer Vision, pp. 471--490. Springer
  (2020)

\bibitem{jain2019two}
Jain, U., Weihs, L., Kolve, E., Rastegari, M., Lazebnik, S., Farhadi, A.,
  Schwing, A.G., Kembhavi, A.: Two body problem: Collaborative visual task
  completion.
\newblock In: Proceedings of the IEEE/CVF Conference on Computer Vision and
  Pattern Recognition, pp. 6689--6699 (2019)

\bibitem{kipf2016semi}
Kipf, T.N., Welling, M.: Semi-supervised classification with graph
  convolutional networks.
\newblock arXiv preprint arXiv:1609.02907  (2016)

\bibitem{ai2thor}
Kolve, E., Mottaghi, R., Han, W., VanderBilt, E., Weihs, L., Herrasti, A.,
  Gordon, D., Zhu, Y., Gupta, A., Farhadi, A.: {AI2-THOR: An Interactive 3D
  Environment for Visual AI}.
\newblock arXiv  (2017)

\bibitem{kurenkov2020semantic}
Kurenkov, A., Mart{\'\i}n-Mart{\'\i}n, R., Ichnowski, J., Goldberg, K.,
  Savarese, S.: Semantic and geometric modeling with neural message passing in
  3d scene graphs for hierarchical mechanical search.
\newblock arXiv preprint arXiv:2012.04060  (2020)

\bibitem{igibson}
Li, C., Xia, F., Mart\'in-Mart\'in, R., Lingelbach, M., Srivastava, S., Shen,
  B., Vainio, K., Gokmen, C., Dharan, G., Jain, T., Kurenkov, A., Liu, K.,
  Gweon, H., Wu, J., Fei-Fei, L., Savarese, S.: igibson 2.0: Object-centric
  simulation for robot learning of everyday household tasks (2021)

\bibitem{liu2020when2com}
Liu, Y.C., Tian, J., Glaser, N., Kira, Z.: When2com: multi-agent perception via
  communication graph grouping.
\newblock In: Proceedings of the IEEE/CVF Conference on Computer Vision and
  Pattern Recognition, pp. 4106--4115 (2020)

\bibitem{liu2020who2com}
Liu, Y.C., Tian, J., Ma, C.Y., Glaser, N., Kuo, C.W., Kira, Z.: Who2com:
  Collaborative perception via learnable handshake communication.
\newblock In: 2020 IEEE International Conference on Robotics and Automation
  (ICRA), pp. 6876--6883. IEEE (2020)

\bibitem{lowe2017multi}
Lowe, R., Wu, Y., Tamar, A., Harb, J., Abbeel, P., Mordatch, I.: Multi-agent
  actor-critic for mixed cooperative-competitive environments.
\newblock arXiv preprint arXiv:1706.02275  (2017)

\bibitem{nachum2019multi}
Nachum, O., Ahn, M., Ponte, H., Gu, S., Kumar, V.: Multi-agent manipulation via
  locomotion using hierarchical sim2real.
\newblock arXiv preprint arXiv:1908.05224  (2019)

\bibitem{puig2018virtualhome}
Puig, X., Ra, K., Boben, M., Li, J., Wang, T., Fidler, S., Torralba, A.:
  Virtualhome: Simulating household activities via programs.
\newblock In: Proceedings of the IEEE Conference on Computer Vision and Pattern
  Recognition, pp. 8494--8502 (2018)

\bibitem{watchandhelp}
Puig, X., Shu, T., Li, S., Wang, Z., Liao, Y.H., Tenenbaum, J.B., Fidler, S.,
  Torralba, A.: Watch-and-help: A challenge for social perception and
  human-{\{}ai{\}} collaboration.
\newblock In: International Conference on Learning Representations (2021).
\newblock \urlprefix\url{https://openreview.net/forum?id=w_7JMpGZRh0}

\bibitem{samvelyan19smac}
Samvelyan, M., Rashid, T., de~Witt, C.S., Farquhar, G., Nardelli, N., Rudner,
  T.G.J., Hung, C.M., Torr, P.H.S., Foerster, J., Whiteson, S.: {The}
  {StarCraft} {Multi}-{Agent} {Challenge}.
\newblock CoRR \textbf{abs/1902.04043} (2019)

\bibitem{habitat}
Savva, M., Kadian, A., Maksymets, O., Zhao, Y., Wijmans, E., Jain, B., Straub,
  J., Liu, J., Koltun, V., Malik, J., Parikh, D., Batra, D.: Habitat: A
  platform for embodied ai research (2019)

\bibitem{alfred}
Shridhar, M., Thomason, J., Gordon, D., Bisk, Y., Han, W., Mottaghi, R.,
  Zettlemoyer, L., Fox, D.: {ALFRED: A Benchmark for Interpreting Grounded
  Instructions for Everyday Tasks}.
\newblock In: The IEEE Conference on Computer Vision and Pattern Recognition
  (CVPR) (2020).
\newblock \urlprefix\url{https://arxiv.org/abs/1912.01734}

\bibitem{tan2020multi}
Tan, S., Xiang, W., Liu, H., Guo, D., Sun, F.: Multi-agent embodied question
  answering in interactive environments.
\newblock In: Computer Vision--ECCV 2020: 16th European Conference, Glasgow,
  UK, August 23--28, 2020, Proceedings, Part XIII 16, pp. 663--678. Springer
  (2020)

\bibitem{wang2021collaborative}
Wang, H., Wang, W., Zhu, X., Dai, J., Wang, L.: Collaborative visual
  navigation.
\newblock arXiv preprint arXiv:2107.01151  (2021)

\bibitem{house3d}
Wu, Y., Wu, Y., Gkioxari, G., Tian, Y.: Building generalizable agents with a
  realistic and rich 3d environment.
\newblock arXiv preprint arXiv:1801.02209  (2018)

\bibitem{zhu2021main}
Zhu, F., Hu, S., Zhang, Y., Hong, H., Zhu, Y., Chang, X., Liang, X.: Main: A
  multi-agent indoor navigation benchmark for cooperative learning  (2021)

\end{thebibliography}
